\documentclass{article}

\usepackage[nonatbib, final]{neurips_2021}
\usepackage[backend=biber,style=numeric]{biblatex}
\usepackage[T1]{fontenc}    
\usepackage{hyperref}       
\usepackage{url}            
\usepackage{booktabs}       
\usepackage{amsfonts}       
\usepackage{nicefrac}       
\usepackage{microtype}      
\usepackage{xcolor}
\usepackage{graphicx}
\usepackage[ruled,vlined,noend]{algorithm2e}
\usepackage{amsmath}
\usepackage{amssymb}
\usepackage{color}
\usepackage{listings}
\usepackage{wrapfig}
\usepackage{amsthm}
\usepackage{subcaption}
\usepackage{multirow}
\usepackage{tikz}
\usetikzlibrary{fit, calc,positioning,matrix, decorations.pathreplacing}

\definecolor{mydarkblue}{rgb}{0,0.08,0.45}
\hypersetup{
    colorlinks=true,
    linkcolor=mydarkblue,
    citecolor=mydarkblue,
    filecolor=mydarkblue,
    urlcolor=mydarkblue
}
 
\definecolor{Code}{rgb}{0,0,0} 
\definecolor{Decorators}{rgb}{0.5,0.5,0.5} 
\definecolor{Numbers}{rgb}{0.5,0,0} 
\definecolor{MatchingBrackets}{rgb}{0.25,0.5,0.5} 
\definecolor{Keywords}{rgb}{0,0,1} 
\definecolor{self}{rgb}{0,0,0} 
\definecolor{Strings}{rgb}{0,0.63,0} 
\definecolor{Comments}{rgb}{0,0.63,1} 
\definecolor{Backquotes}{rgb}{0,0,0} 
\definecolor{Classname}{rgb}{0,0,0} 
\definecolor{FunctionName}{rgb}{0,0,0} 
\definecolor{Operators}{rgb}{0,0,0} 
\definecolor{Background}{rgb}{0.98,0.98,0.98} 
\lstdefinelanguage{Python}{ 
numbers=left, 
numberstyle=\scriptsize, 
numbersep=1em, 
xleftmargin=1em, 
framextopmargin=2em, 
framexbottommargin=2em, 
showspaces=false, 
showtabs=false, 
showstringspaces=false, 
frame=l, 
tabsize=4, 
basicstyle=\small,
backgroundcolor=\color{Background}, 
commentstyle=\color{Comments}\slshape, 
stringstyle=\color{Strings}, 
morecomment=[s][\color{Strings}]{"""}{"""}, 
morecomment=[s][\color{Strings}]{'''}{'''}, 
morekeywords={import,from,class,def,for,while,if,is,in,elif,else,not,and,or,print,break,continue,return,True,False,None,access,as,,del,except,exec,finally,global,import,lambda,print,raise,try,assert}, 
keywordstyle={\color{Keywords}\bfseries}, 
keywordstyle={[2]\color{Decorators}\slshape}, 
emphstyle={\color{self}\slshape}, 
}  
\lstnewenvironment{python}[1][]{%
  \lstset{language=Python,#1}%
}{}

\title{Equinox: neural networks in JAX via callable PyTrees and filtered transformations}

\author{Patrick Kidger\\
University of Oxford\\
The Alan Turing Insitute\\
\texttt{kidger@\hspace{0.8pt}maths.ox.ac.uk}
\And
Cristian Garcia\\
Quansight\\
\texttt{cgarcia@\hspace{0.8pt}quansight.com}
}

\begin{document}

\maketitle

\begin{abstract}
JAX and PyTorch are two popular Python autodifferentiation frameworks. JAX is based around pure functions and functional programming. PyTorch has popularised the use of an object-oriented (OO) class-based syntax for defining parameterised functions, such as neural networks. That this seems like a fundamental difference means current libraries for building parameterised functions in JAX have either rejected the OO approach entirely (Stax) or have introduced OO-to-functional transformations, multiple new abstractions, and been limited in the extent to which they integrate with JAX (Flax, Haiku, Objax). Either way this OO/functional difference has been a source of tension. Here, we introduce `Equinox', a small neural network library showing how a PyTorch-like class-based approach may be admitted without sacrificing JAX-like functional programming. We provide two main ideas. One: parameterised functions are themselves represented as `PyTrees', which means that the parameterisation of a function is transparent to the JAX framework. Two: we filter a PyTree to isolate just those components that should be treated when transforming (`jit', `grad' or `vmap'-ing) a higher-order function of a parameterised function -- such as a loss function applied to a model. Overall Equinox resolves the above tension without introducing any new programmatic abstractions: only PyTrees and transformations, just as with regular JAX. Equinox is available at \url{https://github.com/patrick-kidger/equinox}.
\end{abstract}

\section{Introduction}
JAX is a popular Python framework for autodifferentiation \cite{jax2018github, jax-autodiff}. It has introduced several popular new abstractions, such as the use of `PyTrees' to represent data, and several `transformations', which are higher-order functions -- the most notable of which are `jit', `grad' and `vmap' -- to manipulate pure functions.

These two concepts are central to programming in JAX.

We will assume familiarity with the terminology and syntax of the Python programming language.

\subsection{PyTrees and transformations}

\paragraph{PyTree} A PyTree is any arbitrarily nested composition of `node' types (dictionaries, lists, and tuples) containing `leaf' types (all other Python types). For example,
\begin{python}
variable = [4, {"key1": 3.0, "key2": True, "key3": object()}]
\end{python}
is a PyTree with structure \texttt{[*, \{"key1": *, "key2": *, "key3": *\}]} and leaves \texttt{4, 3.0, True, object()}.

JAX assumes referential transparency in its treatment of PyTrees -- that is, each PyTree describes a tree, and not a directed acyclic graph (DAG). It is generally a mistake to use the same object in multiple leaves of a tree.

JAX allows custom types to be registered as node types. This will be important for Equinox later.

\paragraph{Transformations}
JAX provides several operations which act on pure functions.

`jit' traces a pure function and records its computation graph. This computation graph is just-in-time (JIT) compiled to hardware-specific code via the XLA intermediate representation. The compiled (pure) function is returned. This is primarily used to improve execution speed.

`grad' traces a pure function, and then applies reverse-mode autodifferentiation to return the gradients of the outputs with respect to its inputs.

`vmap' wraps a pure function, so that each of its constituent operations are repeated across a batch of inputs. This is primarily used to write simpler code: the action of a function may be described on just a single element, and then the framework takes care of the batching. There may be implicit efficiency gains: a matrix-vector multiply, batched over the vector, produces an asymptotically more efficient matrix-matrix multiply.

`jit-grad-vmap' is a common pattern when training neural networks. Having described the forward pass of a neural network model, it is vmap'd over a batch of data, this forward pass is differentiated with respect to the parameters of a model, and then this whole operation is JIT-compiled.
\begin{python}
@jax.jit
@jax.grad
def loss(parameters, features, labels):
    pred_labels = jax.vmap(forward)(parameters, features)
    return jax.numpy.mean((labels - pred_labels) ** 2)
\end{python}
In particular this highlights the composability of the different transformations provided by JAX. The above code snippet assumes that a suitable pure function `forward' has been defined.

JAX also provides several other transformations (`vjp', `jvp', `jacrev', `jacfwd', `linearize' and so on) that we will not discuss here.

\subsection{Object-oriented model building}
A parameterised function may be naturally represented as a class, which encapsulates state whilst defining methods parameterised by that state. This encourages an object-oriented (OO) approach to model building.

See for example the following snippet of PyTorch \cite{pytorch} code, in which the weights and biases of the network parameterise the forward pass of the model. \texttt{torchtyping} is used to provide type annotations for tensor shapes \cite{torchtyping}.
\begin{python}
from torch import nn

class Model(nn.Module):
    def __init__(self, in_size: int, hidden_size: int, out_size: int):
        super().__init__()
        self.weight1 = nn.Parameter(torch.randn(in_size, hidden_size))
        self.weight2 = nn.Parameter(torch.randn(hidden_size, out_size))
     
    def forward(self, x: torchtyping.TensorType["batch", "in_size"]
               ) -> torchtyping.TensorType["batch", "out_size"]:
        x = x @ self.weight1
        x = torch.relu(x)
        x = x @ self.weight2
        return x
\end{python}

The primary attraction here is the elegant syntax for model-building. (Not the ability to mutate instance state -- in principle out-of-place updates suffice for most purposes.)

\subsection{Contributions}\label{section:contributions}
We introduce Equinox. Half tech demo, half neural network library, Equinox demonstrates that a PyTorch-like class-based syntax may be used without sacrificing JAX-like functional programming.

To the best of our knowledge Equinox is the first library of its kind, in that it does not exhibit any of the limitations introduced by previous libraries. Indeed Equinox may be most meaningfully defined by what it does \textit{not} have.
\begin{enumerate}
    \item It does not introduce any new abstractions -- everything is either a PyTree or a transformation.
    \item It does not introduce an OO-to-functional transform, for translating from one paradigm to the other.
    \item It does not place limitations or require special care on integrating with arbitrary JAX code.
    \item It does not introduce any specially-wrapped library-specific versions of `jit'/`grad'/`vmap' needed to work with its parameterised functions.\footnote{Equinox does introduce some wrapped `jit'/`grad' operations, namely `filter\_jit' and `filter\_grad' (Section \ref{section:filtering-pytrees}) but these are a convenience only, and operate on arbitrary PyTrees. That is, the key point is that there is no coupling between custom transformations and the representation of parameterised functions.}
\end{enumerate}

That Equinox does not require these complexities makes it easy to learn, makes it absent of `sharp bits' around which a user must be careful, and requires no special effort to interoperate with other JAX libraries.

Equinox has already seen success. In the short time since Equinox was announced, at least three copycat\footnote{A term we use without negative connotation.} libraries have already appeared, quoting Equinox as a primary source of inspiration and offering variations on its ideas \cite{treex, opax, pax}.

The two main ideas introduced with Equinox are the representation of parameterised functions as data (Section \ref{section:parameterised-functions-as-data}) and filtering PyTrees (Section \ref{section:filtering-pytrees}), the latter being particularly useful around transformations such as `jit' and `grad'.

Equinox is available at \url{https://github.com/patrick-kidger/equinox} and may be installed via \texttt{pip install equinox}. At time of writing it is at version 0.1.1. It is available under the Apache-2.0 license.

\section{Related work}\label{section:prior-work}
Note that here and throughout we distinguish the task of building neural networks (requiring definitions of convolutions, dropout and so on) from the lower-level task of representing and manipulating parameterised functions (which is our main focus).

\paragraph{Init/apply}
Stax \cite{stax} forgoes trying to represent parameterised functions entirely. Instead, a parameterised function $f_\theta$ is decomposed into two pieces: $\theta$ and $(\theta, x) \mapsto f_\theta(x)$. The former is a PyTree of data; the latter is a pure function. This is sometimes known as the `init/apply' approach.


\paragraph{Init/apply wrappers}
Haiku's \cite{haiku2020github} central feature is an OO-to-functional operation. A model is built using OO syntax, and then \texttt{haiku.transform} is used to translate between paradigms. The return value is a pair of init/apply functions. However multiple variations are needed to handle all cases (\texttt{haiku.transform\_with\_state}, \texttt{haiku.multi\_transform}, \texttt{haiku.multi\_transform\_with\_state}). Additionally, native JAX operations do not always work within the OO paradigm (only after the OO-to-functional transform), requiring wrapped \texttt{haiku.jit} (and so on) operations to function correctly.

Flax \cite{flax2020github} provides a class-based syntax for representing parameterised functions. Around JAX API boundaries it again operates in init/apply style, and again substantial complexity is introduced. This includes new abstractions like custom parameter groups (\texttt{flax.core.variables.Variable}), wrapped versions of JAX transformations (\texttt{flax.linen.vmap}, \texttt{flax.linen.jit}), and even wrapped versions of other JAX operations (\texttt{flax.linen.scan}).

\paragraph{Complete wrappers}
Objax \cite{objax2020github} offers an OO-based syntax that forgoes almost all interaction with native JAX, in favour of providing a wrapped API for all relevant operations. This involves introducing several new abstractions (variables, modules). Such a `framework-oriented' approach also limits compatibility with third-party JAX libraries.

Objax emphasises (although Flax, Haiku behave similarly) the use of a `Module system'. That is, parameterised functions are represented via a `Module', which is a privileged type with special behaviour. This is contrast to Equinox, later, in which `Module's will be PyTrees like any other.

\paragraph{Classes as PyTrees}
Representing classes as data is a precursor to a class-based syntax for representing parameterised functions as data. In particular, to use a class-based syntax just to group together related information. Flax provides this through
\texttt{flax.struct} and Chex \cite{chex2020github} provides this through \texttt{chex.dataclass}.

\paragraph{Parameterised functions as PyTrees}
\texttt{jax.tree\_util.Partial} represents a parameterised function as data, and may be found within the main JAX library itself. This is philosophically very similar to Equinox -- after instantiation it is a callable PyTree. What it lacks is a convenient class-based syntax for readability and composability.

\paragraph{In other frameworks/languages}
Some notable points of reference exist beyond JAX.

Flux.jl \cite{Flux.jl-2018, innes:2018} in the Julia language \cite{Julia-2017} is philosophically similar to Equinox. It offers a class-based PyTree-like system to define parameterised functions, which it refers to as `functors'. (In reference to the functional programming / category-theoretic notion known in JAX as \texttt{jax.tree\_map}.) Swift for TensorFlow \cite{swift4tf} also does something similar.

There is interest in introducing JAX-like transformations (`grad', `vmap') to the PyTorch framework, via the functorch project \cite{functorch2021}. Just like Haiku this includes an OO-to-functional transform \texttt{functorch.make\_functional} for transforming the PyTorch Module system into pure functions.

\section{Parameterised functions as data}\label{section:parameterised-functions-as-data}

Equinox represents parameterised functions as class instances. These instances are immutable, for consistency with JAX's functional programming principles. (And parameters updates occur out-of-place.)

A parameterised function $f_\theta$ is represented as an instance of a class, with the class corresponding to the function family $\{f_\theta\,|\,\theta \in \Theta \}$. A separate class is defined for each function family.

Every such class is registered as a custom PyTree node type. That is, JAX is told how to (de)serialise class instances into a standardised format.

Class instance state is used to represent the parameterisation $\theta$. Methods may then be used to define the forward operation. Unbound methods define the pure unparameterised function $(\theta, x) \mapsto f_\theta(x)$. Bound methods (with \texttt{self} already passed) define the pure parameterised function $f_\theta$.

That class instances are also PyTrees crucially means that this state is transparent to JAX. That is, given some higher-order pure function $g$ for which $g(f_\theta)$ is defined, then we may for example differentiate $g$ with respect to $f_\theta$ to compute $\tfrac{\mathrm{d} g}{\mathrm{d} f}(f_\theta)$. Indeed this is a typical use-case, such as differentiating a loss function $g(f_\theta, x, y) = (y - f_\theta(x))^2$ with respect to $f_\theta$.

The essential ideas may be summarised in the following code.
\begin{python}
import jax
import jax.numpy as jnp

class Adder:
    def __init__(self, parameter: float):
        self.parameter = parameter
    
    @jax.jit
    def __call__(self, x: jnp.ndarray) -> jnp.ndarray:
        return x + self.parameter
    def tree_flatten(self):
        return (self.parameter,), None
        
    @classmethod
    def tree_unflatten(cls, _, parameter):
        return Adder(parameter[0])

jax.tree_util.register_pytree_node_class(Adder)

@jax.grad
def higher_order_func(adder: Adder, data: jnp.ndarray):
    return adder(data)
\end{python}

The gradient returned by \texttt{higher\_order\_func} will itself be an instance of \texttt{Adder}, whose \texttt{parameter} attribute will be the gradient of the \texttt{parameter} attribute used in the forward pass.

Note how we may JIT-compile \texttt{Adder.\_\_call\_\_}, which is a pure function when including the \texttt{self} argument. This JIT compilation is included by way of demonstration. This is unlike previous libraries: arbitrary JAX transformations (`jit') may safely be used anywhere in the forward pass of the model.

In practice the (de)serialisation and registering as a PyTree node may be neatly handled through a traditional subclassing syntax:
\begin{python}
class Adder(equinox.Module):
    parameter: float
    
    @jax.jit
    def __call__(self, x: jnp.ndarray) -> jnp.ndarray:
        return x + self.parameter
\end{python}

The simplicity of this approach belies the complexities that have been overcome. Indeed we recall our list of things Equinox has \textit{not} required, relative to previous libraries (Section \ref{section:contributions}, Section \ref{section:prior-work}). We sometimes refer to this approach as a `callable PyTree'.

It is actually of some surprise to us that this idea appears to be new to JAX. (Or at least the idea that it is sufficient to build fully-featured models around.) As noted in Section \ref{section:prior-work}, the idea already exists in Julia (Flux.jl functors) and similar ideas already exist in JAX (\texttt{jax.tree\_util.Partial}).

A larger self-contained example is available in Appendix \ref{appendix:example}.

\section{Filtering}\label{section:filtering-pytrees}
There is (only) one foible. The example of Section \ref{section:parameterised-functions-as-data} had one key simplicity; the parameters of \texttt{Adder} -- in practice it had only a single one -- were of a type understood by JAX. More specifically, all of the leaves of the PyTree were of types for which JAX had differentiation rules.

This cannot be true in general. We may wish to parameterise a function by arbitrary Python types, unknown to JAX. More generally we may wish to JIT or differentiate a model with respect to only some of its parameters. Typical examples of this include:
\begin{enumerate}
    \item The activation function in an feedforward network may be an arbitrary Python callable.
    \item It is typical to want to JIT a forward pass with respect to all JAX arrays, but only to differentiate with respect to all floating-point JAX arrays.
    \item It may be desirable to leave some parameters of a model `frozen', and differentiate only with respect to the others.
\end{enumerate}

This foible is handled through `filtering', as in the following example.
\begin{python}
@jax.grad
def loss(parameters, static, data):
    model = equinox.combine(parameters, static)
    return model(data)

model = equinox.nn.MLP(...)
data = ...
parameters, static = equinox.partition(model, equinox.is_array)
loss(parameters, static, data)
\end{python}
A PyTree is partitioned into two pieces on one side of an API boundary -- corresponding to those pieces that should/shouldn't be differentiated -- and then reconstituted again on the other side. Each piece (\texttt{parameters}, \texttt{static}) is another instance of the model-as-PyTree, containing just the leaves that should be treated in the appropriate way and with dummy values on the others.

In practice such partitioning occurs almost exclusively across API boundaries described by JAX transformations. For simplicity (only), Equinox provides wrappers for common use cases.
\begin{python}
@equinox.filter_grad
def loss(model, data):
    return model(data)

model = equinox.nn.MLP(...)
data = ...
loss(model, data)
\end{python}
We emphasise that such `filtered transformations' are entirely unlike the specially-wrapped transformations often introduced in previous libraries, needed to handle the library-specific notion of a parameterised function. There is no coupling between \texttt{filter\_grad} and the model: the former operates on arbitrary PyTrees, whilst the latter is a PyTree like any other.

\section{Further topics}
\paragraph{Design goals} Equinox has two key design goals. One: a very strong regularisation towards simplicity, in particular in the implementation of Equinox itself. Equinox introduces no new abstractions on top of regular JAX: everything is just PyTrees and transformations on PyTrees. 
Two: that using Equinox should feel like (and be fully compatible with) the main JAX library itself. Our litmus test for whether to add a feature is `could this plausibly be in the main JAX library?'.

\paragraph{Location of filtering} Filtering takes place at the call site of a (`jit'/`grad'/`vmap') transformation, and is not embedded into the PyTree structure. For example this is unlike PyTorch (and other JAX libraries) for which whether a tensor participates in autodifferentiation is metadata specified by a \texttt{requires\_grad} flag attached to the tensor itself. This is in-line with native JAX, and furthermore ensures compatibility with any other transformations introduced into JAX at a later date.

\paragraph{Lack of module system} Equinox Modules are PyTrees like any other, and are never special-cased. This is unlike PyTorch, and other JAX libraries, which for example often feature a `module map' analogous to \texttt{jax.tree\_map}.

\paragraph{Bound methods are PyTrees too} Section \ref{section:parameterised-functions-as-data} discusses letting class instances be PyTrees, so that \text{\_\_call\_\_} is a pure parameterised function transparent to JAX. We go further and treat all bound methods as PyTrees, via Python's descriptor protocol. That is, the bound method is a PyTree whose single child subtree is its implicit \texttt{self} parameter.

\paragraph{Managing state} `State' is used to refer to parameters updated other than through gradient descent, such as batch normalisation statistics. This may be handled by mutating the model in-place as desired, and then returning the updated model out of any JAX API boundaries \cite{treex}.

\paragraph{Neural network library} The focus of Equinox is the representation and manipulation of parameterised functions as data. Neural networks are of course a major use case for parameterised functions, so for convenience (and as proof of its efficacy) Equinox also includes a small \texttt{equinox.nn} library.

\section{Conclusion}
We have introduced Equinox, which demonstrates how we may use a PyTorch-like class-based API for building models (parameterised functions) without sacrificing JAX-style functional programming.

\begin{ack}
PK was supported by the EPSRC grant EP/L015811/1 and by the Alan Turing Institute under the EPSRC grant EP/N510129/1.
\end{ack}

\printbibliography
\appendix
\clearpage
\section{Example}\label{appendix:example}
The following is example code for defining a custom parameterised function in Equinox.

It demonstrates how to compose one parameterised function with another (the linear layers) and how to define new parameters directly (the bias).

It additionally demonstrates the use of filtered transformations, to handle the fact that some of the leaves of the model-as-PyTree are nondifferentiable/non-JIT-able.

\begin{python}
import equinox as eqx
import jax
import jax.nn as jnn
import jax.numpy as jnp
import jax.random as jrandom

class MyModule(eqx.Module):
    # Specify the module's attributes;
    layers: list          # nested list of submodules
    activation: callable  # arbitrary Python object
    bias: jnp.ndarray     # parameter
    
    # And how to initialise them;
    def __init__(self, key):
        key1, key2 = jrandom.split(key)
        self.layers = [eqx.nn.Linear(2, 8, key=key1),
                       eqx.nn.Linear(8, 2, key=key2)]
        self.activation = jnn.relu
        self.bias = jnp.ones(2)

    # And the forward pass of the model.
    def __call__(self, x):
        for layer in self.layers[:-1]:
            x = self.activation(layer(x))
        return self.layers[-1](x) + self.bias

@eqx.filter_jit
@eqx.filter_grad
def loss(model, x, y):
    pred_y = jax.vmap(model)(x)
    return jnp.mean((y - pred_y) ** 2)

x_key, y_key, model_key = jrandom.split(jrandom.PRNGKey(0), 3)
x, y = jrandom.normal(x_key, (100, 2)), jrandom.normal(y_key, (100, 2))
model = MyModule(model_key)
grads = loss(model, x, y)
\end{python}
\end{document}